\documentclass[letterpaper, 10 pt, conference]{ieeeconf}  

\IEEEoverridecommandlockouts                              

\usepackage{graphics} 
\usepackage{epsfig} 
\usepackage{amsmath} 
\usepackage{amssymb}  
\usepackage{threeparttable}

\usepackage{graphicx}
\usepackage{subfigure}

\usepackage{verbatim}
\usepackage{epstopdf}
\usepackage{tabularx}
\usepackage{cite}
\usepackage{booktabs}
\usepackage{multirow}
\usepackage{color}
\usepackage{bm}

\title{\LARGE \bf
Towards Two-view 6D Object Pose Estimation: A Comparative Study on Fusion Strategy
}

\author{Jun Wu$^{1}$, Lilu Liu$^{1}$, Yue Wang$^{1}$ and Rong Xiong$^{1}$
\thanks{$^{1}$Jun Wu, Lilu Liu, Yue Wang, and Rong Xiong are with the State Key Laboratory
of Industrial Control Technology and Institue of Cyber-Systems and Control,
Zhejiang University, Zhejiang, China.}
\thanks{Corresponding author,
        {\tt\small wangyue@iipc.zju.edu.cn},
        Co-corresponding author,
        {\tt\small rxiong@zju.edu.cn}}.
}

\begin{document}

\maketitle
\thispagestyle{empty}
\pagestyle{empty}

\begin{abstract}

Current RGB-based 6D object pose estimation
methods have achieved noticeable performance 
on datasets and real world applications. 
However, predicting 6D pose
from single 2D image features is susceptible to disturbance from
changing of environment and textureless or resemblant object
surfaces. Hence, RGB-based methods generally achieve less
competitive results than RGBD-based methods, which deploy
both image features and 3D structure features. To narrow down
this performance gap, this paper proposes a framework for 6D
object pose estimation that learns implicit 3D information from
2 RGB images. Combining the learned 3D information and 2D
image features, we establish more stable correspondence between
the scene and the object models. To seek for the methods
best utilizing 3D information from RGB inputs, we conduct an
investigation on three different approaches, including Early-
Fusion, Mid-Fusion, and Late-Fusion. We ascertain the Mid-
Fusion approach is the best approach to restore the most precise
3D keypoints useful for object pose estimation. The experiments
show that our method outperforms state-of-the-art RGB-based
methods, and achieves comparable results with RGBD-based
methods.

\end{abstract}

\section{Introduction}

6D object pose estimation aspires to estimate the rotation and translation of 
interested objects with regard to certain canonical coordinates. Accurate object 
pose estimation is the key to many real-world applications, such as robotic manipulation,
augmented reality, and human-robot interactions. This is a challenging problem due to the 
variety of objects appearance, occlusions between objects, and clutter in the scene. 

Based on the sensors they adopt, current model-based object pose estimation methods 
can be roughly categorized into two classes: 
RGB-based methods \cite{xiang2017posecnn}\cite{park2019pix2pose}\cite{zakharov2019dpod}
and RGBD-based methods \cite{wang2019densefusion}\cite{wada2020morefusion}\cite{hua2021rede}. 
Though the scale is a known quantity in model-based object pose estimation problem, 
previous researches have shown that the performance of RGB-based methods  
are generally less competitive than the RGBD-based methods\cite{shugurov2021dpodv2}, 
mostly due to the lacking of 3D structure information.
Since an image is the projection of the object under certain lighting condition 
and observation angle, predicting object pose with only RGB inputs could be
affected by low resolutions, ill observation pose, changing of environment, 
and textureless or resemblant surface, thus involves more ambiguities. 
On the other hand, RGBD-based methods usually integrate 3D structures with 2D image 
features to estimate object poses\cite{shi2021stablepose}\cite{he2020pvn3d}. 
The factors impacting one kind of input modality could 
barely impact the other, thus this integration of two modalities of input features 
creates more robust prediction results. 

\begin{figure}
        \centering
        \includegraphics[width=0.45\textwidth]{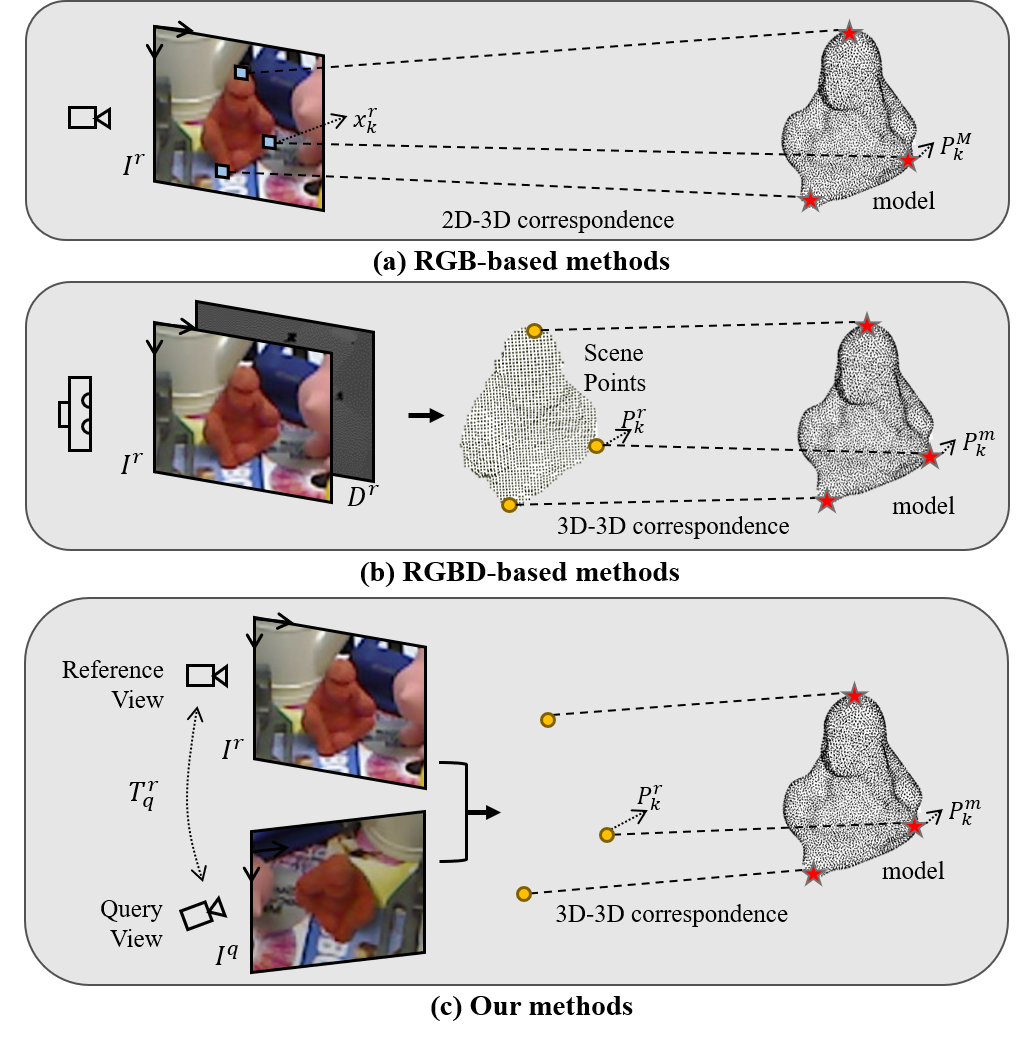}
        \caption{\textbf{Illustration of our ideas.} 
        We show the general pipelines of RGB-based, RGBD-based methods and our methods. 
        RGB-based methods take RGB image as input, establish correspondence between 
        image features and object models, then solve for the pose. While RGBD-based 
        methods deploy image features and point cloud features to build the correspondence,
        which enhances the robustness of estimation.
        Our motivation is to narrow down the performance gap between RGB-based 
        methods and RGBD-based methods, by proposing a framework combining image features
        from 2 input images to learn the implicit 3D information. }
        \vspace{-0.55cm}
        \label{fig:idea}
\end{figure}

Therefore, a reasonable approach to enhance the pose estimation accuracy 
of RGB-based methods would be learning 3D geometric information from the RGB inputs.
Though 2D images lack awareness of 3D geometric information in nature, 
two or more frames of such images combined together implicitly bring out the 
depth information, as has been verified in many stereo or multi-view stereo tasks 
\cite{chang2018pyramid}\cite{li2019stereo}\cite{yao2018mvsnet}.  
Thus, we opt to learn the implicit 3D information from two or more RGB 
images to enhance the accuracy in estimating object poses. 
Luckily, making two or more observations of the scene with a certain relative 
transformation is commonly accessible for robots, rendering the applicability 
of this pipeline.



In this paper, we investigate 3 different approaches to learn the 3D information from 
2 input images, including fusing images to restore depth map (Early-Fusion),
fusing 2D keypoints predicted from each image to restore 3D keypoints (Late-Fusion), 
and fusing 2D feature maps to restore characterized 3D space points (Mid-Fusion).
We analyze all approaches on real word datasets and discover that the Mid-Fusion approach
is able to restore the most precise 3D information useful for object pose estimation 
task. And by comparing to other state-of-the-art methods, we prove that combining 
2D image features with learned 3D information effectively enhance the pose 
estimation accuracy.

To this end, the major contributions of this paper are as follows:


\begin{itemize}


\item To narrow down the performance gap between RGB-based methods and 
RGBD-based methods, we propose a framework for 6D object pose estimation, 
which learns the implicit 3D information from two RGB images 
to solve the pose. 
\item To seek for the method best utilizing 3D information useful for object 
pose estimation task, we conduct an investigation on 3 different approaches,
including Early-Fusion, Mid-Fusion, and Late-Fusion, and show that the 
Mid-Fusion approach yields more precise 3D information. 
\item We compare our method to state-of-the-art methods on benchmarks.
Our results show that integrating 3D information with 2D features effectively 
enhance the object pose estimation accuracy.   

\end{itemize}


\section{Related Works}


\textbf{RGB-based Object Pose Estimation.}
Many object pose estimation methods take RGB images as input, and tackle this problem by learning the 
correspondence between 2D image features and 3D CAD models 
\cite{park2019pix2pose}\cite{rad2017bb8}\cite{zakharov2019dpod}\cite{hodan2020epos}\cite{manhardt2019explaining}.  
\cite{tekin2018real} and \cite{luo20203d} employs CNN 
to predict the 3D bounding box corners of the object in the image, and associate the 
corners with those in 3D CAD models to solve the pose. Since the corners are artificial, 
the estimation results are not satisfactory. 
To use more reliable correspondence, PVNet \cite{peng2019pvnet} selects keypoints from the object's model,
and train a CNN to predict the vertex from every pixel to those keypoints then vote with confidence. 
Besides keypoints, HybridPose \cite{song2020hybridpose} employs edge vectors and symmetry
correspondence to enrich the feature space for better estimation. 
Beyond establishing correspondence between 2D features and 3D models, some methods seek to 
directly learn the rationale between 2D features and 6D poses with neural 
networks\cite{xiang2017posecnn}\cite{hu2020single}. 
Furthermore, \cite{wang2021gdr} and \cite{di2021so} explicitly learn the dense 2D-3D 
correspondence between images and canonical models, then regress poses 
by a Patch-PnP network.

\textbf{RGBD-based Object Pose Estimation.}
To further improve the estimation accuracy, another pipeline, 
RGBD-based methods additionally deploy depth information 
\cite{wang2019densefusion}\cite{wada2020morefusion}\cite{saadi2021optimizing}\cite{he2021ffb6d}. 
Early research \cite{hinterstoisser2012model} 
\cite{hinterstoisser2011multimodal} compose contour vectors
from RGB image and surface normal vectors from depth image, and estimate the pose by 
template matching. PVN3D \cite{he2020pvn3d} uses a neural network to separately extract 
image and point features, then encode them pointwisely to vote for 3D keypoints, and solve 
the pose with 3D-3D correspondence. REDE \cite{hua2021rede} and L6DNet \cite{gonzalez2021l6dnet} 
both encode multimodal features, and apply a 3D registration algorithm to solve the pose.
Because of the extra depth information, RGBD-based methods generally achieve better 
results than RGB-based methods. 

\textbf{Multi-view Stereo Object Pose Estimation.}
Recently, some methods deploy multiframe observations to further refine pose 
estimation results in a optimization framework
\cite{collet2010efficient}\cite{collet2011moped}\cite{labbe2020cosypose}\cite{fu2021multi}\cite{shugurov2021dpodv2}.
They formulate a global optimization function to refine the object poses predicted from 
each observation by minimizing the reprojection loss or average distance loss.
This backend framework is generally adaptive to work with other frontend 
pose estimation methods, including our proposed one. 
\cite{liu2020keypose} is the method most similar to ours, which proposes a pose 
estimation framework for transparent objects. 
They fuse the stereo RGB inputs to restore dense depth map, assigning depth value 
to the predicted 2D keypoints, then solve the 3D-3D registration problem. 
Our proposed Early-Fusion approach shares the same idea with them, and will be discussed 
later in the experiments.

\textbf{Multi-view Stereo Scene Reconstruction.}
Restoring scene geometry from multiple overlapping images is a problem 
widely studied.
Early Methods decouples the complex multi-view problem into relatively small 
problems with only one reference and a few source images at a time 
\cite{2008Using}\cite{Engin2011Efficient}.
Based on plane-sweep stereo, many recent learning-based 
methods\cite{yao2019recurrent}\cite{luo2019p}\cite{yang2020cost}
build cost volumes with warped 2D image features from multiple observations, 
regularize them with 3D CNNs and regress the depth. 
Cas-MVSNet\cite{2020Cascade} proposes a cascade cost volume mechanism based on feature pyramids
and estimates the depth in a coarse-to-fine manner to reduce the computation resources.
We follow them to build our Early-Fusion pipeline to restore the depth and evaluate
its capability in restoring useful 3D keypoints for pose estimation.


\begin{figure*}
        \begin{center}
        \includegraphics[width=0.95\textwidth]{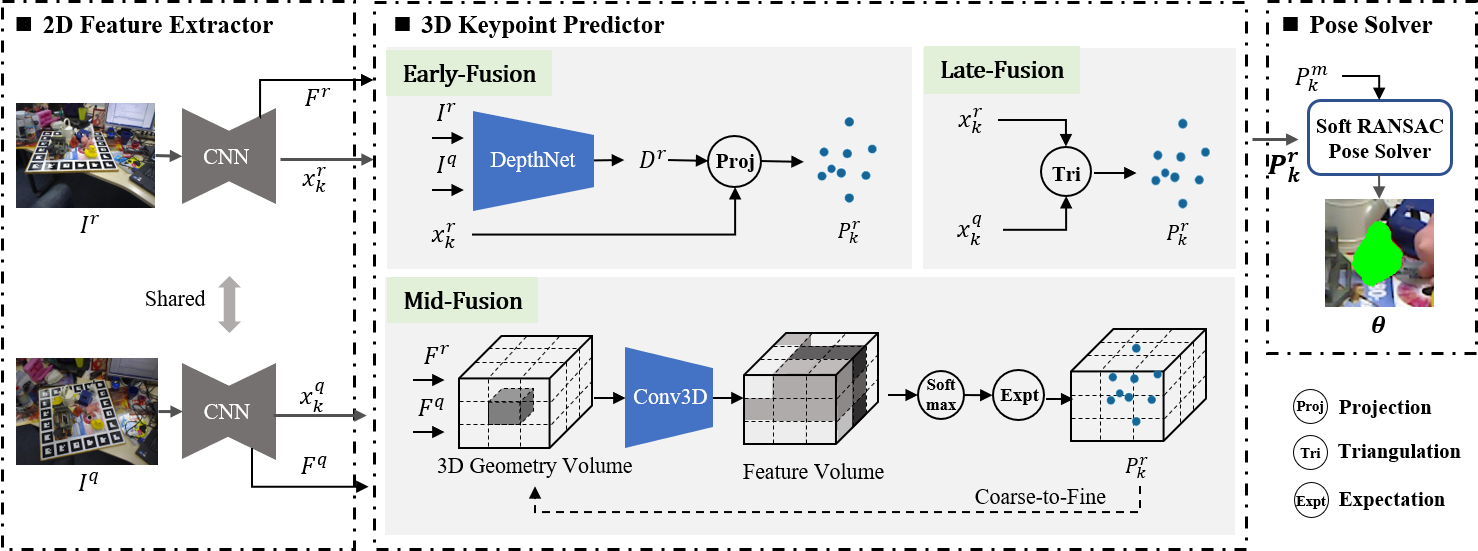}
        \caption{\textbf{Methods Overview.} 
        We show the overall pipeline of our proposed methods. 
        Taking the input reference
        image $I^{r}$ and query image $I^{q}$, we extract 2D image feature $F^{r}$ and $F^{q}$
        with a shared multi-scale feature extractor network.
        Then, 3 different approaches are developed to predict 3D keypoints $P_{k}^{r}$, including 
        Early-Fusion, Mid-Fusion, and Late-Fusion. 
        The Early-Fusion approach fuses the two input images to predict depth map $D^{r}$ 
        via a DepthNet,
        thus the predicted 2D keypoints $x_{k}^{r}$ are projected to $P_{k}^{r}$. 
        The Late-Fusion approach fuses $x_{k}^{r}$ and $x_{k}^{q}$ by directly
        triangulating them to get $P_{k}^{r}$.
        The Mid-Fusion approach fuse the two feature maps to restore a characterized
        3D geometry volume $V_{r}$, then predicts $P_{k}^{r}$ with 3D CNNs.
        Notice that for a fair comparison, the 2D keypoints $x_{k}^{r}$ are  
        predicted by an additional head of the shared feature extractor.
        Last, $P_{k}^{r}$ are used to solve the pose in a closed form with model keypoints
        $P_{k}^{m}$.}
        \vspace{-0.55cm}
        \label{fig:framework}
        \end{center}
\end{figure*}

\section{Method}

In this section, we introduce our overall pipeline to combine two images for object pose 
estimation, as illustrated in Fig.~\ref{fig:framework}. 
Taking the input reference
image $I^{r}$ and query image $I^{q}$, we extract 2D image feature $F^{r}$ and $F^{q}$
with a shared feature extractor network.
Then, 3 different approaches are developed to predict 3D keypoints $P_{k}^{r}$, including 
Early-Fusion, Mid-Fusion, and Late-Fusion. 
The Early-Fusion approach fuse the two input images to obtain depth map $D^{r}$,
thus the predicted 2D keypoints $x_{k}^{r}$ are projected to $P_{k}^{r}$. 
The Mid-Fusion approach fuse the two feature maps to restore a characterized
3D geometry volume $V^{r}$, then predicts $P_{k}^{r}$ with 3D CNNs.
And the Late-Fusion approach fuses $x_{k}^{r}$ and $x_{k}^{q}$ directly to 
get $P_{k}^{r}$.
Notice that for a fair comparison, the 2D keypoints $x_{k}^{r}$ are  
predicted by an additional head of the shared feature extractor.
Last, $P_{k}^{r}$ are used to solve the pose in a closed form with model keypoints
$P_{k}^{m}$. 

\subsection{2D Feature Extraction} 

For a fair comparison, we adopt a shared feature extraction network for all our approaches.
Since feature extraction is not the focus of this paper, we follow \cite{peng2019pvnet}
to build a multi-scale convolutional neural network, to extract features
at multiple resolutions, and predicts 2D keypoints $\{x_{k}^{r}, x_{k}^{q}\}$ for 
Early-Fusion pipeline and Late-Fusion pipeline. 


\subsection{3D Keypoints Prediction} 

With the goal of learning useful 3D information from two input RGB images, we explore
3 different approaches to predict the 3D keypoints, including Early-Fusion, Mid-Fusion,
and Late-Fusion. Here we present how we develop these approaches separately.

\textbf{Early-Fusion Approach:} The most intuitive way to learn 3D information from 
input images is to directly recover depth from them. 
This approach takes advantage of the abundant off-the-shelf methods in 
the multi-view stereo depth estimation field, thus is rather facile to implement.
Without loss of generality, we follow \cite{2020Cascade} to build a cascade 
of 3D geometry volumes from warped 2D image features, regularize them with 
independent 3D CNN blocks and regress the depth map. 
Then, for a fair comparison with the other two approaches, we predict 2D keypoints 
from the input reference image following \cite{peng2019pvnet}. 
The extracted image features are fed into a vector prediction head to predict the 
vectors from each pixel to its predicted keypoints pixels, then a RANSAC voting 
mechanism is applied to decide the mean and covariance of each keypoint. 
Last, the 3D keypoints are obtained by assigning depth values to the predicted 2D 
keypoints. 

\textbf{Late-Fusion Approach:} Another nature approach to learn 3D information 
is to fuse the predicted 2D keypoints from both inputs into 3D keypoints. 
Likewise, we follow \cite{peng2019pvnet} to predict 2D keypoint for equitable 
contrast. Then, we triangulate these 2D keypoints to 3D with their relative camera
poses by OpenCV library algorithm.


\textbf{Mid-Fusion Approach:} Despite the two straightforward approaches, 
we take a deeper look at the essence of keypoint prediction process.
Predicting keypoints is to seek the location where the 
features best match the target keypoint previously chosen in the object models.
Hence, the distinctiveness of the features is significant. 
Considering this, the Early-Fusion and the Late-Fusion approaches both rely 
solely on the 2D image features to predict, and the 3D information is only used 
to project the 2D keypoints to 3D. 
Therefore, we consider this Mid-Fusion approach to combine 2D image features and 
3D information together to predict the keypoints. 

As shown in Fig.~\ref{fig:framework}, we build a 3D geometry volume from the 
two image feature maps, regularize it with 3D CNNs, then reduce the divergence 
between the feature field and the local keypoint heatmaps to predict 3D keypoints.

To build the 3D geometry volume, we divide the interested 3D space 
into regular 3D grids of size $(H_{v},W_{v},D_{v})$, and the size of each grid is 
$(h_{v},w_{v},d_{v})$. The axes of the grid coordinates are centralized in an initial guess 
and parallel to the reference camera coordinates. 
We create a many-to-one projection from 3D grid space to the image space by 
known camera intrinsic parameters.
The grid space is also projected to the query image space by the relative
camera pose between the two views.
Then we assign the features extracted before of the 2D point to its related 3D point.
Points projected outside the image are assigned to initial feature values. 
At each 3D point, we simply concatenate the two feature vectors from two inputs 
to keep as much information as possible. 

Then, to predict 3D keypoints, we regularize this 3D volume with 3D CNNs to 
obtain a field representing the distribution of target keypoint locations. 
This distribution is learned by reducing the divergence between the 3D feature 
field and the local target keypoint heatmaps. 
The local keypoint heatmaps are defined as 
\begin{equation}
Q_{i}^{k}(p|P_{ki}^{m},\theta) \sim \mathcal{N}(\theta \cdot P_{ki}^{m},\sigma_{i})
\end{equation}
where $\{P_{ki}^{m}\}_{i=1}^N$ refers to the model keypoints, $\theta$ is the target 
pose, and $\sigma_{i}$ is the hyperparameters. This distribution represents the 
local heatmap we expect to be highlighted as target keypoints locations.
And the feature field is defined as
\begin{equation}
Q_{i}^{v}(p|V(\cdot)) = \sum_{j\in \Omega} w_{j}V([p_{j}])
\end{equation}
where $\{w_{j}\}_{j\in \Omega}$ are the trilinear interpolation coefficients, 
$[p_{j}]$ are the 8 neighbour grids of the conditioned position $p$, 
and $V(\cdot)$ is the value operation in the feature field. This distribution 
describes how likely a position in the field is to be the keypoints.
To minimal the divergence of these two distributions, we adopt a Kullback-Leibler
divergence loss
\begin{equation}
Loss_{KL} = D_{KL}(Q^{v}\|Q^{k}) = -\sum_{i=1}^{N} Q_{i}^{v}log(\frac{Q_{i}^{k}}{Q_{i}^{v}})
\end{equation}  

Last, we predict the 3D keypoint by calculating the expectation of the distribution 
in the 3D volume.
Besides, we employ a coarse-to-fine strategy to further enhance the prediction accuracy.

\subsection{Soft RANSAC Solver}

Given the predicted 3D keypoints $\{\hat{P}_{ki}\}_{i=1}^N$ and the model keypoints 
$\{P_{ki}^{m}\}_{i=1}^N$, we then solve the object pose by minimizing the distance 
between predicted points and model points. The optimization problem is 
\begin{equation}
\hat{\theta} = \arg\min_{\theta}\|\theta \cdot P_{ki}^{m}-\hat{P}_{ki}\|_{2}
\end{equation}        
where $\theta$ represents the 6D object pose.

Though this optimization problem can be solved by SVD in closed form, it is 
easy to be disturbed by outliers. 
Therefore, we propose a soft RANSAC solver to softly count the inliers to 
solve a more robust pose. Taking the predicted keypoints $\{\hat{P}_{ki}\}_{i=1}^N$, 
we calculate all possible poses with every 3 points by SVD, which brings us a 
set of pose hypothesis ${\{\theta_{k}\}_{k=1}^{{C_{N}^{3}}}}$. 
For each hypothesis, we evaluate the distance for every predicted and 
model keypoints under the hypothesis pose
\begin{equation}
d_{k,i} = \|\theta_{k} \cdot P_{ki}^{m}-\hat{P}_{ki}\|_{2}
\end{equation}       
Then we sum up the soft inlier numbers for each hypothesis to evaluate the pose, and
softly aggregate all the hypothesis into a final pose with a regularized scores
\begin{equation}
\hat{\theta} = \sum_{k}^{K} \theta_{k} \cdot\sum_{i}^{N}sigmoid(\gamma_{1}(-d_{k,i}+\gamma_{2}))
\end{equation}


This solver is shared among all the 3 proposed approaches. 

For the differentiability in the entire process in Mid-Fusion approach, 
the pose error could be end-to-end back-propagated to train the 3D networks.
The loss to evaluate the predicted pose is 
\begin{equation}
Loss_{pose} = \|\hat{t}-t\|_{2}+\alpha \|\hat{R}R^{T}-I\|_{F}
\end{equation}
In total, the Mid-Fusion approach is trained with joint losses
\begin{equation}
Loss = \sum_{j}\beta_{1}Loss_{pose_{j}}+\beta_{2}Loss_{kpt_{j}} +\beta_{3} Loss_{KL_{j}}
\end{equation}
where $j=\{0,1\}$ refers to the coarse and fine levels, 
and $\{\beta_{1}, \beta{2}, \beta{3}\}$ are hyperparameters.


\begin{figure}
        \begin{center}
        \includegraphics[width=0.48\textwidth]{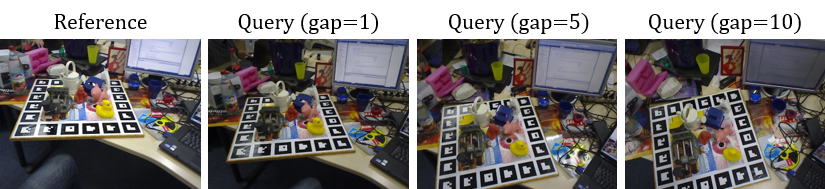}
        \caption{\textbf{Example of query images under different pairing gaps.} 
        The reference image and query image are paired up by a fixed gap, in 
        which $gap=1$ means pairing two images with a gap of one frame, and 
        so on. Generally larger gaps lead to farther views. }
        \vspace{-0.55cm}
        \label{fig:gaps}
        \end{center}
\end{figure}

\section{Experiments}

In this section, we evaluate our method by experiments to answer two questions: 
(1) Which of the proposed fusing approaches can recover the most precise and useful 3D 
information from 2 RGB images for object pose estimation? 
(2) Combined with 2D image features and 3D information, can our method 
enhance the pose estimation accuracy?
To answer (1), we assess and compare the precision of the predicted 
3D keypoints of the proposed approaches.
To answer (2), we compare the estimation results between our method 
and other state-of-the-art methods on public benchmark LineMOD dataset\cite{hinterstoisser2011multimodal} 
and Occlusion LineMOD dataset\cite{brachmann2014learning}.

\begin{table}[htbp]
        \centering
        \caption{evaluation on 3d keypoints prediction. 
        \vspace{-0.3cm}
                ($3D\mbox{-}Keypoint\mbox{-}Distance/m$)}
                \renewcommand{\arraystretch}{1.2}        
                \begin{threeparttable}
                \begin{tabular}{lcccc} 
                        \toprule
                        & Early-Fusion & Mid-Fusion & Late-Fusion & REDE\cite{hua2021rede} \\ 
                        \midrule
                        $gap=1$  & 0.044 & \textbf{0.024} & 0.079 &  \\ 
                        $gap=5$  & 0.031 & \textbf{0.009} & 0.014 & \\ 
                        $gap=10$  & 0.031 & \textbf{0.007} & 0.011 & \\
                        \midrule
                        mean & 0.035 & \textbf{0.013} & 0.035 & 0.024 \\
                        \bottomrule
                \end{tabular}
                \end{threeparttable}
        \label{tab:kpt-error}%
        \vspace{-0.55cm}
\end{table}

\begin{figure}[htbp]
        \centering
        \subfigure[ape]{
                \includegraphics[width=4.12cm]{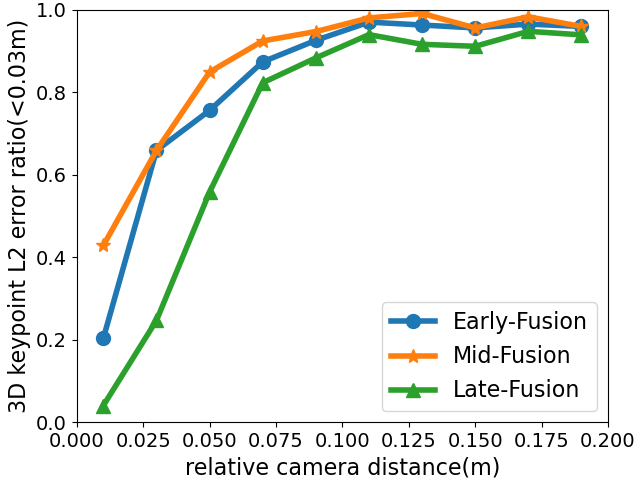}}
        \subfigure[can]{
                \includegraphics[width=4.12cm]{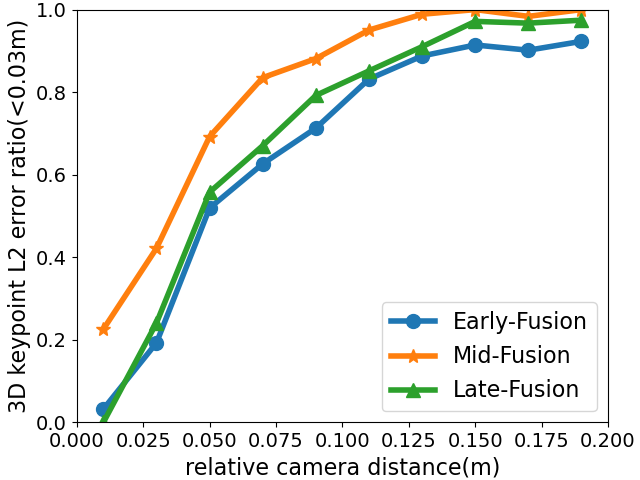}}
        \subfigure[cat]{
                \includegraphics[width=4.12cm]{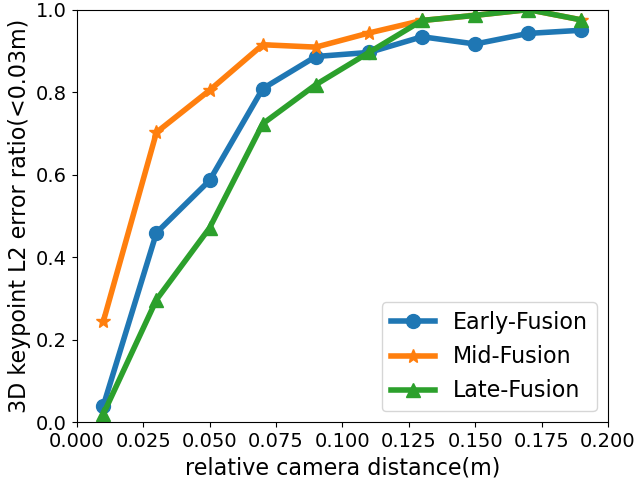}}
        \subfigure[glue]{
                \includegraphics[width=4.12cm]{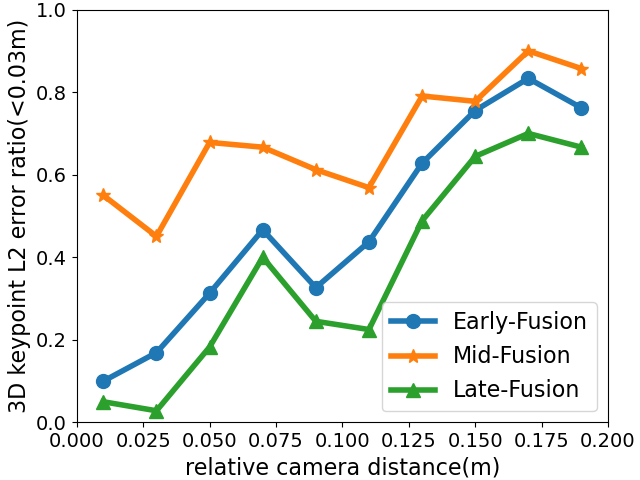}}
        \subfigure[eggbox]{
                \includegraphics[width=4.12cm]{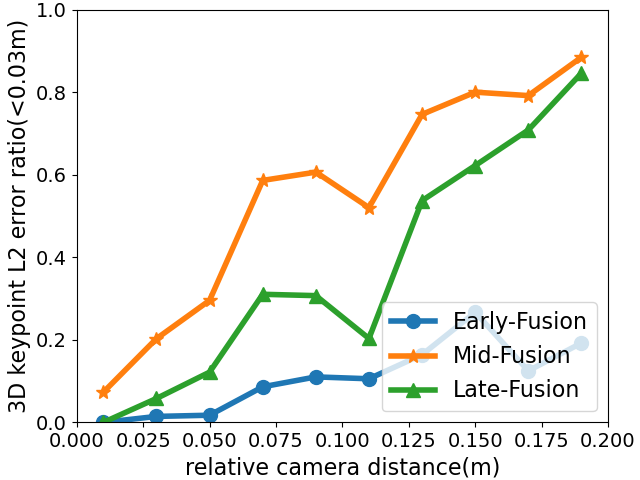}}
        \subfigure[iron]{
                \includegraphics[width=4.12cm]{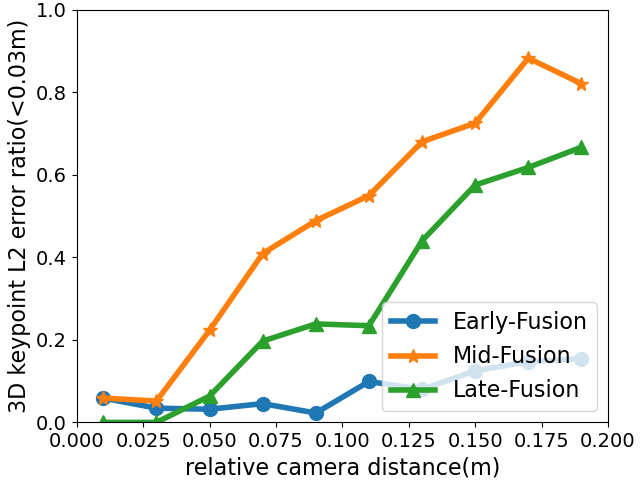}}
        \caption{\textbf{3D Keypoint Recall ($<0.03m$) with respect to relative camera distances.}
                Each point represents a ratio of the numbers of keypoints with an 
                error less than $0.03m$ to the number of all keypoints under certain 
                relative camera distances. The relative camera distances are computed 
                as the distances between two focal points between the two cameras.
                The Mid-Fusion approach achieves better results in most of the 
                camera distances, especially in small parallax.}
        \vspace{-0.5cm}
        \label{fig:kpt-error}
\end{figure}

\subsection{Dataset}

LineMOD dataset\cite{hinterstoisser2011multimodal} is a widely used benchmark 
for object 6D pose estimation tasks. 
It contains 13 objects with varieties of textures and structures.
For a fair comparison, we follow \cite{peng2019pvnet} \cite{hua2021rede} to split it
into train, valid and test data. For each object, about 180 images are for training 
and more than 1000 images are for testing. Also, we follow \cite{peng2019pvnet} to 
further create 10000 images by "Cut and Paste" strategy to train all the objects.

Occlusion LineMOD\cite{brachmann2014learning} dataset shares the same objects with 
LineMOD, but additionally annotates a subset of images with serious occlusion and clutters. 
It contains 8 different objects, and 1214 images. All images are used for testing 
with models trained on LineMOD dataset.

\subsection{Metrics}

We follow convention to use ADD \cite{hinterstoisser2012model}
and ADD-S \cite{xiang2017posecnn} as evaluation metrics.
Given an object model of $M$ points, 
ADD metric evaluates the average distance between the model points transformed with 
predicted and ground truth pose respectively 
\begin{equation}
ADD = \frac{1}{M}\sum_{i=1}^{M}\|\theta p_{i}-\hat{\theta} p_{i}\|_{2}
\end{equation}

While ADD-S metric calculates the average distance between the closest points, which 
is used for evaluating symmetric object
\begin{equation}
ADD\mbox{-}S= \frac{1}{M}\sum_{i=1}^{M}\min_{j\in M}\|\theta p_{j}-\hat{\theta} p_{i}\|_{2}
\end{equation}

An estimation is regarded successful if the ADD(-S) is less than 
$10\%$ of the object's diameter. 

Besides, to evaluate the quality of restored 3D information, we apply 3D keypoint 
Euclidean distance between predicted 3D keypoints and ground truth 3D keypoints.

\begin{figure}[htbp]
        \centering
        \includegraphics[width=0.48\textwidth]{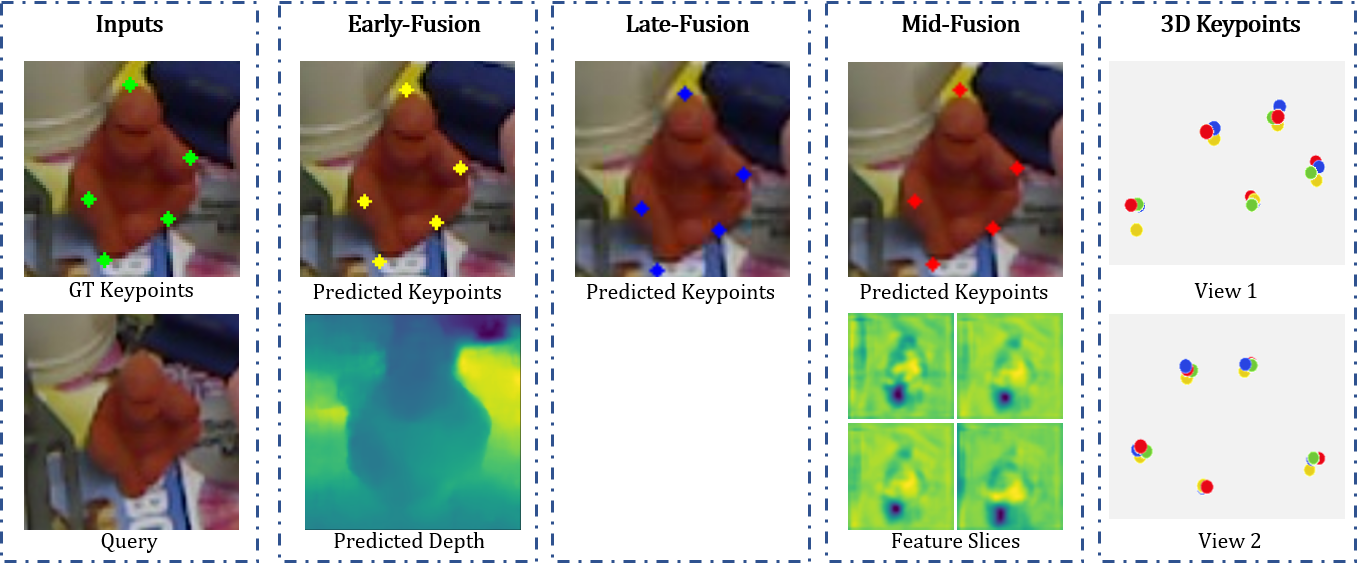}
        \caption{\textbf{Visualization of keypoint prediction results from different fusion approach.} 
        From left to right: input image pairs with ground truth 3D keypoints
        (green), predicted keypoints from Early-Fusion 
        (yellow) and the predicted depth map, 
        predicted keypoints from Late-Fusion (blue),
        predicted keypoints from Mid-Fusion (red) and four 
        slices of the feature volume, 
        and all the ground truth and predicted 3D keypoints in two different views
        with respective colors.
        }
        \vspace{-0.5cm}
        \label{fig:middle}
\end{figure}

\begin{table*}[htbp]
        \centering
        \caption{Performance comparison on LineMOD (ADD(-S)$<0.1d$).}
        \vspace{-0.3cm}
            \begin{threeparttable}
            \begin{tabular}{l|cccccccc|cc}
                \toprule
                & \multicolumn{8}{c|}{RGB} & \multicolumn{2}{c}{RGBD} \\
                \midrule
                & Pix2Pose & PVNet & DPOD & CDPN & HybridPose & SO-Pose & GDR-Net & \textbf{Ours} & DenseFusion & REDE  \\
                & \cite{park2019pix2pose} & \cite{peng2019pvnet} & \cite{zakharov2019dpod} & \cite{li2019cdpn} & \cite{song2020hybridpose} & \cite{di2021so} & \cite{wang2021gdr} & & \cite{wang2019densefusion} & \cite{hua2021rede} \\
                \midrule
                \midrule
                ape & 58.1 & 43.6 & 53.3 & 64.4 & 63.1 & & & \textbf{94.9} & 92.3 & 96.3 \\
                benchvise & 91.0 & 99.9 & 95.3 & 97.8 & \textbf{99.9} & & & \textbf{99.9} & 93.2 & 98.9 \\
                cam & 60.9 & 86.9 & 90.4 & 91.7 & 90.4 & & & \textbf{93.7} & 94.4 & 99.8 \\
                can & 84.4 & 95.5 & 94.1 & 95.9 & 98.5 & & & \textbf{99.0} & 93.1 & 99.4 \\
                cat & 64.0 & 79.3 & 60.4 & 83.8 & 89.4 & & & \textbf{98.3} & 96.5 & 99.2  \\
                driller & 76.3 & 96.4 & 97.7 & 96.2 & 98.5 & & & \textbf{99.1} & 87.0 & 99.3 \\
                duck & 43.8 & 52.6 & 66.0 & 66.8 & 65.0 & & & \textbf{96.0} & 92.3 & 96.2 \\
                eggbox* & 96.8 & 99.1 & 99.7 & 99.7 & \textbf{100.0} & & & 99.3 & 99.8 & 100.0 \\
                glue* & 79.4 & 95.7 & 93.8 & \textbf{99.6} & 98.8 & & & 99.0 & 100.0 & 100.0 \\
                holepuncher & 74.8 & 81.9 & 65.8 & 85.8 & 89.7 & & & \textbf{94.5} & 86.9 & 99.2 \\
                iron & 83.4 & 98.9 & 99.8 & 97.9 & \textbf{100.0} & & & 98.8 & 97.0 & 99.9 \\
                lamp & 82.0 & 99.3 & 88.1 & 97.9 & \textbf{99.5} & & & 98.4 & 95.3 & 99.5 \\
                phone & 45.0 & 92.4 & 74.2 & 90.1 & 94.9 & & & \textbf{96.5} & 92.8 & 98.9 \\
                \midrule
                mean & 72.4 & 86.3 & 83.0 & 89.9 & 91.3 & 96.0 & 94.1 & \textbf{97.5} & 94.3 & 99.0 \\
                \bottomrule
            \end{tabular}%
            \begin{tablenotes}
                \item *denotes symmetric objects.
            \end{tablenotes}
            \end{threeparttable}
        \label{tab:lm}%
        \vspace{-0.5cm}
\end{table*}%

\subsection{Implementation Details}

In implementation, we follow \cite{peng2019pvnet} to select 9 keypoints for 
every object, and perform the same data augmentation. 
Every two adjacent frames are paired up for training, and every two frames 
with a certain gap are paired up for testing. 
The relative camera poses are obtained from the dataset.

In the Early-Fusion approach, we finetune the pretrained models provided by 
\cite{2020Cascade} on our datasets with an initial learning rate of $0.0001$ 
and a stepwise decay strategy. 
The depth range is set according to the 
depth range of the dataset, and the depth resolution is set to $3.5mm$.

In the Mid-Fusion approach, the keypoint prediction network contains three 3D 
convolutional layers all with one 3D BatchNorm layer and 1 ReLU layer respectively, 
1 output 3D convolutional layer, and finally 1 LogSoftmax layer.
The 3D geometric volume is set in range
$[-0.3,0.3]\times [-0.3,0.3]\times [-0.3,0.3]$(meters) with grid size of $0.01m$
in the coarse level. 
While the range of the fine volume is dependent on the diameter of each object 
and its grid size is set to $0.005m$.



We run all our experiments on a machine equipped with an 
Intel(R) Xeon(R) Silver 4216 CPU at 2.10GHz, and an NVIDIA GeForce RTX 3090 GPU.

\subsection{Evaluation on 3D Keypoint Precision}

To validate the ability of our method to restore 3D information from the input RGB
images, we train all the networks in the 3 proposed approaches on 
LineMOD dataset, then analyze and compare the quality of the predicted 3D object 
keypoints in terms of average 3D Keypoint Distance metric. 
During testing, we adopt a fixed gapping strategy to pair up images as inputs for 
our proposed method.
Since the dataset is collected with handheld cameras, the relative camera poses
between each frame are diverse even with fixed gaps. 
But generally a larger gap still leads to two farther views, as the example 
in Fig.~\ref{fig:gaps} shows.
Thus, to analyze the effect of pairing gaps, 
the evaluation is conducted under different gaps ($\{1,5,10\}$),
in which $gap=1$ means pairing two frames with a gap of one frame, and so on.

As can be seen in Table.~\ref{tab:kpt-error}, with the increase of gap frames, 
the 3D keypoints prediction precision of the proposed methods are all enhanced.
Among all the 3 approaches, the Mid-Fusion approach achieves the best  
prediction results, with an average keypoint error of $0.007m$ in $gap=10$. 
The Late-Fusion approach is the second-best in $gap=5$ and $gap=10$ occasions.
While the Early-Fusion approach obtains the worst results in all pairing settings.
We also compare the keypoints predicted by our method and those predicted by 
a SOTA RGBD-based method \cite{hua2021rede}. 
Table.~\ref{tab:kpt-error} shows that, compared to \cite{hua2021rede} ($0.024m$),
the Early-Fusion approach achieves slightly worse results, 
and the Late-Fusion approach gets better results with $gap=5$ ($0.014m$) and $gap=10$ ($0.011m$).
What's more, the Mid-Fusion approach achieves the same results even under $gap=1$, 
and better results with larger gaps.

To further verify the performance of the 3 proposed approaches under different 
parallax, we draw an accuracy ratio curve under different relative camera distances
in 4 different objects (ape, can, cat, and glue), as shown in Fig.~\ref{fig:kpt-error}.
Each point in the figure represents a ratio of the numbers of keypoints with an 
error less than $0.03m$ to the number of all keypoints under certain 
relative camera distances. The relative camera distances are computed 
as the distances between two focal points between the two cameras.
The Mid-Fusion approach achieves the best results in most of the 
camera distances, especially in small ones.
It exhibits the robustness and superiority of the Mid-Fusion approach in 
different situations. 

Fig.~\ref{fig:middle} shows a visualized result comparison among 
the 3 proposed approaches. 
The 3D predicted keypoints are projected to the reference image by camera 
parameters, as well as shown in 2 different views in 3D space.
Also, the depth map predicted by the Early-Fusion approach and the feature slices
learned by the Mid-Fusion approach are exhibited.
As can be seen, though some predictions show good performance when projected to 2D image 
plane, their errors are more clear when observed in the 3D space, such as 
the top keypoint predicted by the Early-Fusion approach.

To this end, we argue that among all the 3 proposed methods, the Mid-Fusion 
approach achieves the best prediction results under different situations, 
which shows its ability to restore the most useful 3D information for object 
pose estimation from two input RGB images.
Therefore, we adopt the Mid-Fusion approach as the keypoint predictor to conduct 
experiments compared to other state-of-the-art methods.

\begin{table*}[htbp]
        \centering
        \caption{Performance comparison on Occlusion LineMOD (ADD(-S)$<0.1d$).}
        \vspace{-0.3cm}
            \begin{threeparttable}
            \begin{tabular}{l|ccccccc|ccc}
                \toprule
                & \multicolumn{7}{c|}{RGB} & \multicolumn{3}{c}{RGBD} \\
                \midrule
                & PoseCNN & PVNet & DPOD & HybridPose & SO-Pose & GDR-Net & \textbf{Ours} & PVN3D & REDE & FFB6D \\
                & \cite{xiang2017posecnn} & \cite{peng2019pvnet} & \cite{zakharov2019dpod} & \cite{song2020hybridpose} & \cite{di2021so} & \cite{wang2021gdr} & & \cite{he2020pvn3d} & \cite{hua2021rede} & \cite{he2021ffb6d} \\
                \midrule
                \midrule
                ape & 9.6 & 15.8 & & 20.9 & 48.4 & 46.8 & \textbf{49.2} & 33.9 & 55.9 & 47.2 \\
                can & 45.2 & 63.3 & & 75.3 & 85.8 & \textbf{90.8} & 89.3 & 88.6 & 87.9 & 85.2 \\
                cat & 0.93 & 16.7 & & 24.9 & 32.7 & \textbf{40.5} & 37.7 & 39.1 & 37.5 & 45.7 \\
                driller & 41.4 & 65.7 & & 70.2 & 77.4 & 82.6 & \textbf{90.1} & 78.4 & 75.3 & 81.4 \\
                duck & 19.6 & 25.2 & & 27.9 & 48.9 & 46.9 & \textbf{57.8} & 41.9 & 48.5 & 53.9 \\
                eggbox* & 22.0 & 50.2 & & 52.4 & 52.4 & 54.2 & \textbf{57.3} & 80.9 & 73.3 & 70.2 \\
                glue* & 38.5 & 49.6 & & 53.8 & \textbf{78.3} & 75.8 & 71.6 & 68.1 & 74.8 & 60.1 \\
                holepuncher & 22.1 & 39.7 & & 54.2 & 75.3 & 60.1 & \textbf{80.0} & 74.7 & 61.9 & 85.9 \\
                \midrule
                mean & 24.9 & 40.8 & 47.3 & 47.5 & 62.3 & 62.2 & \textbf{66.6} & 63.2 & 64.2 & 66.2 \\
                \bottomrule
            \end{tabular}%
            \begin{tablenotes}
                \item *denotes symmetric objects.
            \end{tablenotes}
            \end{threeparttable}
        \label{tab:occ}%
        \vspace{-0.5cm}
\end{table*}%

\subsection{Evaluation on Object Pose Estimation}

To validate the enhancement of our method after combining 2D image features and 
learned 3D information, we evaluate the performance in LineMOD dataset and 
Occlusion LineMOD dataset and compete with state-of-the-art RGB-based methods 
and RGBD-based methods. In this section, all our methods are evaluated with $gap=10$
pairing strategy.

Table.~\ref{tab:lm} displays the performance on LineMOD dataset. 
Best ADD(-S) recalls for each object are in bold, excluding RGBD-based methods.
Our method excels all other RGB-based methods with a mean ADD recall of $97.5$,
and most of the methods in 9 object classes.  
Also, our method performs slightly better than the RGBD-based method \cite{wang2019densefusion} \
in 10 object classes and in the mean recall metric. 
Both indicate the advantage of combining 2D image features and 3D information 
together for pose estimation. 
Nonetheless, though predicts more accurate 3D keypoints than \cite{hua2021rede},
our method performs worse than \cite{hua2021rede} in pose estimation. 
We consider the reason lies in the pose solver they adopt. 
In \cite{hua2021rede}, after predicting sparse 3D keypoints, they deploy 
a mechanism called minimal solver bank to use dense real scene 
points recovered from the depth sensors to softly aggregate all the possible poses 
calculated from 3D registration, which is indeed an exhaustive RANSAC algorithm. 
The solver is robust enough to remedy the uncertainty brought by less accurate keypoints. 
Yet out Mid-Fusion approach is unable to deploy such an algorithm due to the 
lack of dense 3D points, while the Early-Fusion approach restores dense points
with certain errors which will bring more disturbance to the solver. 

Pose estimation under serious occlusion situations is a difficult task,
especially for RGB-based methods. 
Table.~\ref{tab:occ} shows the performance on Occlusion LineMOD dataset. 
Best ADD(-S) recalls for each object is in bold, excluding RGBD-based methods.
In this dataset with more occlusion and clutter, the mean ADD(-S) recall of our 
method beats all other RGB-based and RGBD-based methods.
If excluding RGBD-based, our method beats RGB-based methods in 5 object classes. 
Note that we also beat \cite{hua2021rede} in this dataset.
Under the heavily occluded situation, our method not only take advantage of both 
2D image features and learned 3D information, but is also benefited from 
extra observation views. Hence, we have more opportunities to acquire useful information
for keypoint prediction and pose estimation.

\begin{figure*}
        \begin{center}
        \includegraphics[width=0.95\textwidth]{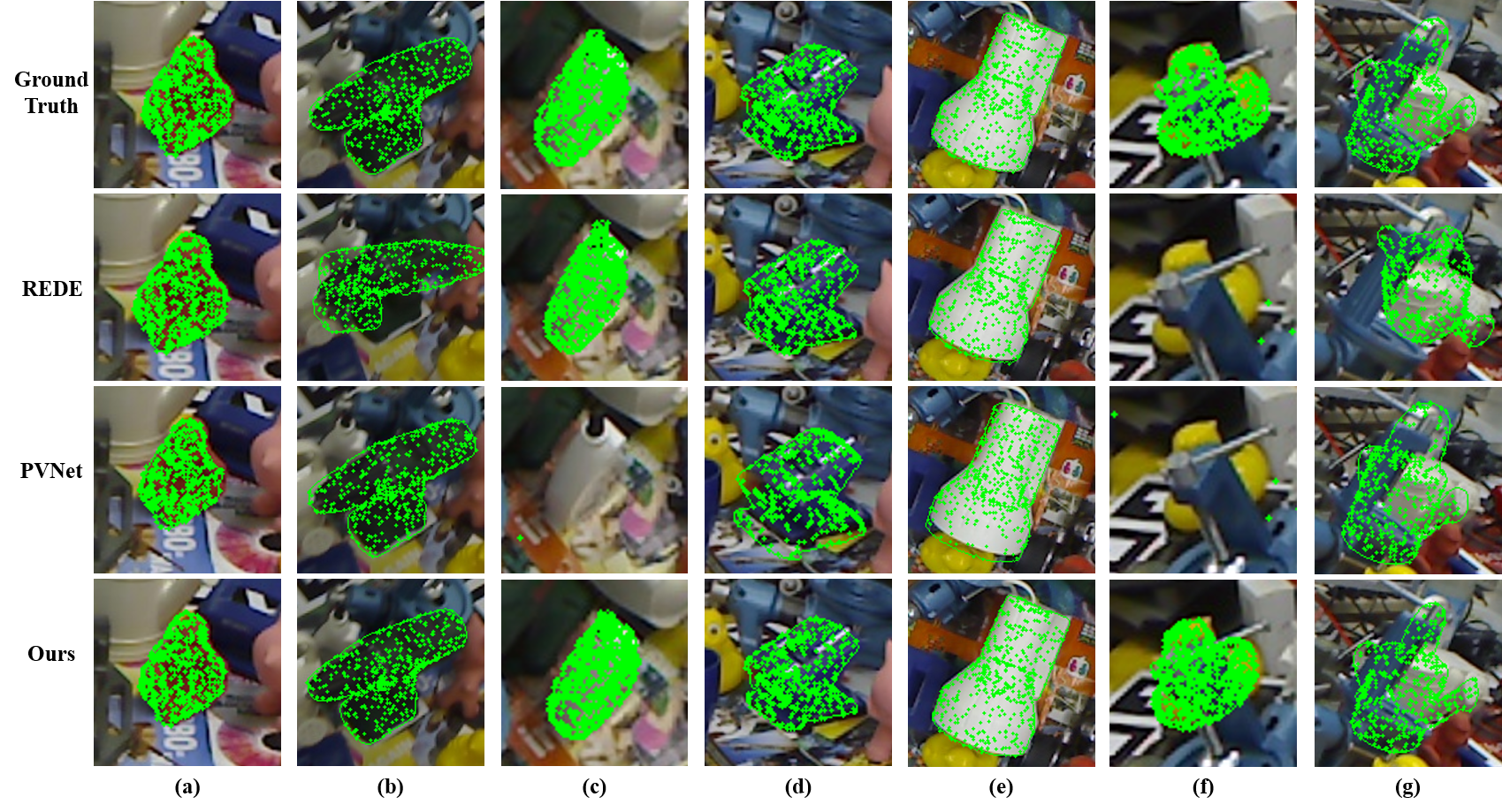}
        \caption{\textbf{Visualization Results Samples.} 
        Some visualization results on LindMOD and Occlusion LineMOD dataset. 
        The pose results are shown by projecting the model points via the predicted
        pose from each method. We also show the contour of the model projection mask.
        All images are cropped for better visualization.
        The cases with few dislocated projected points represent a predicted pose
        with large errors.
        The last row shows our result with Mid-Fusion approach. 
        Compared with RGB-based method PVNet\cite{peng2019pvnet},
        our method achieves more accurate results in textureless surfaces (e),
        bizarre viewing angle (c), and some normal situations (a)(b)(d).
        Also, our methods beats \cite{peng2019pvnet} and RGBD-based method\cite{hua2021rede}
        in occluded occasions (f)(g).}
        \vspace{-0.6cm}
        \label{fig:results}
        \end{center}
\end{figure*}

Some visualization results are shown in Fig.~\ref{fig:results}. 
We project the object CAD model transformed by
the estimated pose and draw the points on the reference view.
All the images are cropped for better visualization.
Compared with RGB-based method PVNet\cite{peng2019pvnet},
our method achieves more accurate results in textureless surfaces,
bizarre viewing angle, and some normal situations.
Also, our method beats both \cite{peng2019pvnet} and RGBD-based method\cite{hua2021rede}
in occluded occasions.

\section{CONCLUSIONS}

In this paper, we propose a framework to learn implicit 3D information from 2 input
RGB images for 6D object pose estimation. In this framework, we conduct an 
investigation on 3 different 3D information learning approaches.
Based on the experiments, we reveal a Mid-Fusion mechanism that fuses 2D image 
features geometrically to 3D space produces the most precise keypoints.
We further show that with this Mid-Fusion mechanism, our framework is capable of 
enhancing the pose estimation performance of RGB-based methods, and achieving 
comparable results with RGBD-based methods. 
However, our performance in less occluded environments are still inferior to 
RGBD-based methods, mostly because of the lack of dense reliable scene points. 
Thus, we plan to tackle the challenge by employing two or more views of RGBD 
inputs to solve the pose in the future.







\bibliographystyle{ieeetr}
\bibliography{reference.bib} 

\begin{thebibliography}{10}

\bibitem{xiang2017posecnn}
Y.~Xiang, T.~Schmidt, V.~Narayanan, and D.~Fox, ``Posecnn: A convolutional
  neural network for 6d object pose estimation in cluttered scenes,'' {\em
  arXiv preprint arXiv:1711.00199}, 2017.

\bibitem{park2019pix2pose}
K.~Park, T.~Patten, and M.~Vincze, ``Pix2pose: Pixel-wise coordinate regression
  of objects for 6d pose estimation,'' in {\em Proceedings of the IEEE/CVF
  International Conference on Computer Vision}, pp.~7668--7677, 2019.

\bibitem{zakharov2019dpod}
S.~Zakharov, I.~Shugurov, and S.~Ilic, ``Dpod: 6d pose object detector and
  refiner,'' in {\em Proceedings of the IEEE/CVF International Conference on
  Computer Vision}, pp.~1941--1950, 2019.

\bibitem{wang2019densefusion}
C.~Wang, D.~Xu, Y.~Zhu, R.~Mart{\'\i}n-Mart{\'\i}n, C.~Lu, L.~Fei-Fei, and
  S.~Savarese, ``Densefusion: 6d object pose estimation by iterative dense
  fusion,'' in {\em Proceedings of the IEEE/CVF conference on computer vision
  and pattern recognition}, pp.~3343--3352, 2019.

\bibitem{wada2020morefusion}
K.~Wada, E.~Sucar, S.~James, D.~Lenton, and A.~J. Davison, ``Morefusion:
  Multi-object reasoning for 6d pose estimation from volumetric fusion,'' in
  {\em Proceedings of the IEEE/CVF conference on computer vision and pattern
  recognition}, pp.~14540--14549, 2020.

\bibitem{hua2021rede}
W.~Hua, Z.~Zhou, J.~Wu, H.~Huang, Y.~Wang, and R.~Xiong, ``Rede: End-to-end
  object 6d pose robust estimation using differentiable outliers elimination,''
  {\em IEEE Robotics and Automation Letters}, vol.~6, no.~2, pp.~2886--2893,
  2021.

\bibitem{shugurov2021dpodv2}
I.~Shugurov, S.~Zakharov, and S.~Ilic, ``Dpodv2: Dense correspondence-based 6
  dof pose estimation,'' {\em IEEE Transactions on Pattern Analysis and Machine
  Intelligence}, 2021.

\bibitem{shi2021stablepose}
Y.~Shi, J.~Huang, X.~Xu, Y.~Zhang, and K.~Xu, ``Stablepose: Learning 6d object
  poses from geometrically stable patches,'' in {\em Proceedings of the
  IEEE/CVF Conference on Computer Vision and Pattern Recognition},
  pp.~15222--15231, 2021.

\bibitem{he2020pvn3d}
Y.~He, W.~Sun, H.~Huang, J.~Liu, H.~Fan, and J.~Sun, ``Pvn3d: A deep point-wise
  3d keypoints voting network for 6dof pose estimation,'' in {\em Proceedings
  of the IEEE/CVF conference on computer vision and pattern recognition},
  pp.~11632--11641, 2020.

\bibitem{chang2018pyramid}
J.-R. Chang and Y.-S. Chen, ``Pyramid stereo matching network,'' in {\em
  Proceedings of the IEEE Conference on Computer Vision and Pattern
  Recognition}, pp.~5410--5418, 2018.

\bibitem{li2019stereo}
P.~Li, X.~Chen, and S.~Shen, ``Stereo r-cnn based 3d object detection for
  autonomous driving,'' in {\em Proceedings of the IEEE/CVF Conference on
  Computer Vision and Pattern Recognition}, pp.~7644--7652, 2019.

\bibitem{yao2018mvsnet}
Y.~Yao, Z.~Luo, S.~Li, T.~Fang, and L.~Quan, ``Mvsnet: Depth inference for
  unstructured multi-view stereo,'' in {\em Proceedings of the European
  Conference on Computer Vision (ECCV)}, pp.~767--783, 2018.

\bibitem{rad2017bb8}
M.~Rad and V.~Lepetit, ``Bb8: A scalable, accurate, robust to partial occlusion
  method for predicting the 3d poses of challenging objects without using
  depth,'' in {\em Proceedings of the IEEE International Conference on Computer
  Vision}, pp.~3828--3836, 2017.

\bibitem{hodan2020epos}
T.~Hodan, D.~Barath, and J.~Matas, ``Epos: Estimating 6d pose of objects with
  symmetries,'' in {\em Proceedings of the IEEE/CVF conference on computer
  vision and pattern recognition}, pp.~11703--11712, 2020.

\bibitem{manhardt2019explaining}
F.~Manhardt, D.~M. Arroyo, C.~Rupprecht, B.~Busam, T.~Birdal, N.~Navab, and
  F.~Tombari, ``Explaining the ambiguity of object detection and 6d pose from
  visual data,'' in {\em Proceedings of the IEEE/CVF International Conference
  on Computer Vision}, pp.~6841--6850, 2019.

\bibitem{tekin2018real}
B.~Tekin, S.~N. Sinha, and P.~Fua, ``Real-time seamless single shot 6d object
  pose prediction,'' in {\em Proceedings of the IEEE Conference on Computer
  Vision and Pattern Recognition}, pp.~292--301, 2018.

\bibitem{luo20203d}
Q.~Luo, H.~Ma, L.~Tang, Y.~Wang, and R.~Xiong, ``3d-ssd: Learning hierarchical
  features from rgb-d images for amodal 3d object detection,'' {\em
  Neurocomputing}, vol.~378, pp.~364--374, 2020.

\bibitem{peng2019pvnet}
S.~Peng, Y.~Liu, Q.~Huang, X.~Zhou, and H.~Bao, ``Pvnet: Pixel-wise voting
  network for 6dof pose estimation,'' in {\em Proceedings of the IEEE/CVF
  Conference on Computer Vision and Pattern Recognition}, pp.~4561--4570, 2019.

\bibitem{song2020hybridpose}
C.~Song, J.~Song, and Q.~Huang, ``Hybridpose: 6d object pose estimation under
  hybrid representations,'' in {\em Proceedings of the IEEE/CVF conference on
  computer vision and pattern recognition}, pp.~431--440, 2020.

\bibitem{hu2020single}
Y.~Hu, P.~Fua, W.~Wang, and M.~Salzmann, ``Single-stage 6d object pose
  estimation,'' in {\em Proceedings of the IEEE/CVF conference on computer
  vision and pattern recognition}, pp.~2930--2939, 2020.

\bibitem{wang2021gdr}
G.~Wang, F.~Manhardt, F.~Tombari, and X.~Ji, ``Gdr-net: Geometry-guided direct
  regression network for monocular 6d object pose estimation,'' in {\em
  Proceedings of the IEEE/CVF Conference on Computer Vision and Pattern
  Recognition}, pp.~16611--16621, 2021.

\bibitem{di2021so}
Y.~Di, F.~Manhardt, G.~Wang, X.~Ji, N.~Navab, and F.~Tombari, ``So-pose:
  Exploiting self-occlusion for direct 6d pose estimation,'' in {\em
  Proceedings of the IEEE/CVF International Conference on Computer Vision},
  pp.~12396--12405, 2021.

\bibitem{saadi2021optimizing}
L.~Saadi, B.~Besbes, S.~Kramm, and A.~Bensrhair, ``Optimizing rgb-d fusion for
  accurate 6dof pose estimation,'' {\em IEEE Robotics and Automation Letters},
  vol.~6, no.~2, pp.~2413--2420, 2021.

\bibitem{he2021ffb6d}
Y.~He, H.~Huang, H.~Fan, Q.~Chen, and J.~Sun, ``Ffb6d: A full flow
  bidirectional fusion network for 6d pose estimation,'' in {\em Proceedings of
  the IEEE/CVF Conference on Computer Vision and Pattern Recognition},
  pp.~3003--3013, 2021.

\bibitem{hinterstoisser2012model}
S.~Hinterstoisser, V.~Lepetit, S.~Ilic, S.~Holzer, G.~Bradski, K.~Konolige, and
  N.~Navab, ``Model based training, detection and pose estimation of
  texture-less 3d objects in heavily cluttered scenes,'' in {\em Asian
  conference on computer vision}, pp.~548--562, Springer, 2012.

\bibitem{hinterstoisser2011multimodal}
S.~Hinterstoisser, S.~Holzer, C.~Cagniart, S.~Ilic, K.~Konolige, N.~Navab, and
  V.~Lepetit, ``Multimodal templates for real-time detection of texture-less
  objects in heavily cluttered scenes,'' in {\em 2011 international conference
  on computer vision}, pp.~858--865, IEEE, 2011.

\bibitem{gonzalez2021l6dnet}
M.~Gonzalez, A.~Kacete, A.~Murienne, and E.~Marchand, ``L6dnet: Light 6 dof
  network for robust and precise object pose estimation with small datasets,''
  {\em IEEE Robotics and Automation Letters}, vol.~6, no.~2, pp.~2914--2921,
  2021.

\bibitem{collet2010efficient}
A.~Collet and S.~S. Srinivasa, ``Efficient multi-view object recognition and
  full pose estimation,'' in {\em 2010 IEEE International Conference on
  Robotics and Automation}, pp.~2050--2055, IEEE, 2010.

\bibitem{collet2011moped}
A.~Collet, M.~Martinez, and S.~S. Srinivasa, ``The moped framework: Object
  recognition and pose estimation for manipulation,'' {\em The international
  journal of robotics research}, vol.~30, no.~10, pp.~1284--1306, 2011.

\bibitem{labbe2020cosypose}
Y.~Labb{\'e}, J.~Carpentier, M.~Aubry, and J.~Sivic, ``Cosypose: Consistent
  multi-view multi-object 6d pose estimation,'' in {\em European Conference on
  Computer Vision}, pp.~574--591, Springer, 2020.

\bibitem{fu2021multi}
J.~Fu, Q.~Huang, K.~Doherty, Y.~Wang, and J.~J. Leonard, ``A multi-hypothesis
  approach to pose ambiguity in object-based slam,'' in {\em 2021 IEEE/RSJ
  International Conference on Intelligent Robots and Systems (IROS)},
  pp.~7639--7646, IEEE, 2021.

\bibitem{liu2020keypose}
X.~Liu, R.~Jonschkowski, A.~Angelova, and K.~Konolige, ``Keypose: Multi-view 3d
  labeling and keypoint estimation for transparent objects,'' in {\em
  Proceedings of the IEEE/CVF conference on computer vision and pattern
  recognition}, pp.~11602--11610, 2020.

\bibitem{2008Using}
N.~D.~F. Campbell, G.~Vogiatzis, C.~Hernández, and R.~Cipolla, ``Using
  multiple hypotheses to improve depth-maps for multi-view stereo,'' in {\em
  European Conference on Computer Vision}, 2008.

\bibitem{Engin2011Efficient}
Engin, Tola, Christoph, Strecha, Pascal, and Fua, ``Efficient large-scale
  multi-view stereo for ultra high-resolution image sets,'' {\em Machine Vision
  and Applications}, vol.~23, no.~5, pp.~903--920, 2011.

\bibitem{yao2019recurrent}
Y.~Yao, Z.~Luo, S.~Li, T.~Shen, T.~Fang, and L.~Quan, ``Recurrent mvsnet for
  high-resolution multi-view stereo depth inference,'' in {\em Proceedings of
  the IEEE/CVF Conference on Computer Vision and Pattern Recognition},
  pp.~5525--5534, 2019.

\bibitem{luo2019p}
K.~Luo, T.~Guan, L.~Ju, H.~Huang, and Y.~Luo, ``P-mvsnet: Learning patch-wise
  matching confidence aggregation for multi-view stereo,'' in {\em Proceedings
  of the IEEE/CVF International Conference on Computer Vision},
  pp.~10452--10461, 2019.

\bibitem{yang2020cost}
J.~Yang, W.~Mao, J.~M. Alvarez, and M.~Liu, ``Cost volume pyramid based depth
  inference for multi-view stereo,'' in {\em Proceedings of the IEEE/CVF
  Conference on Computer Vision and Pattern Recognition}, pp.~4877--4886, 2020.

\bibitem{2020Cascade}
X.~Gu, Z.~Fan, S.~Zhu, Z.~Dai, and P.~Tan, ``Cascade cost volume for
  high-resolution multi-view stereo and stereo matching,'' in {\em 2020
  IEEE/CVF Conference on Computer Vision and Pattern Recognition (CVPR)}, 2020.

\bibitem{brachmann2014learning}
E.~Brachmann, A.~Krull, F.~Michel, S.~Gumhold, J.~Shotton, and C.~Rother,
  ``Learning 6d object pose estimation using 3d object coordinates,'' in {\em
  European conference on computer vision}, pp.~536--551, Springer, 2014.

\bibitem{li2019cdpn}
Z.~Li, G.~Wang, and X.~Ji, ``Cdpn: Coordinates-based disentangled pose network
  for real-time rgb-based 6-dof object pose estimation,'' in {\em Proceedings
  of the IEEE/CVF International Conference on Computer Vision}, pp.~7678--7687,
  2019.

\end{thebibliography}


\end{document}